\documentclass[letterpaper, 10 pt, conference]{ieeeconf}  

\IEEEoverridecommandlockouts                              

\title{\bf ARDuP: Active Region Video Diffusion for Universal Policies}

\author{Shuaiyi Huang$^{1*}$, Mara Levy$^{1}$, Zhenyu Jiang$^{2}$, Anima Anandkumar$^{3}$,\\ Yuke Zhu$^{2,4}$, Linxi Fan$^{4}$, De-An Huang$^{4}$, Abhinav Shrivastava$^{1}$
\thanks{*This work was partially done during Shuaiyi's internship at NVIDIA.}
\thanks{$^{1}$Shuaiyi Huang, Mara Levy, and Abhinav Shrivastava are with University of Maryland, College Park. \{huangshy, mlevy, abhinav2\}@umd.edu}%
\thanks{$^{2}$Zhenyu Jiang and Yuke Zhu are with University of Texas at Austin. \{zhenyu, yukez\}@cs.utexas.edu}%
\thanks{$^{3}$Anima Anandkumar is with California Institute of Technology. anima@caltech.edu}%
\thanks{$^{4}$Yuke Zhu, Linxi Fan, and De-An Huang are with NVIDIA Corporation. \{yukez, linxif, deahuang\}@nvidia.com }%
}

\usepackage{pifont}
\usepackage{bbm}
\usepackage{caption}
\usepackage{subcaption}
\let\labelindent\relax
\usepackage{enumitem} 
\usepackage{xspace}
\usepackage{makecell}
\usepackage{graphicx}
\usepackage{booktabs}
\usepackage{multirow}
\usepackage{tikz}
\usepackage{amsmath}
\usepackage{makecell}

\usepackage[pagebackref,breaklinks,colorlinks]{hyperref}
\usepackage[capitalize]{cleveref}
\usepackage{rotating}

\newcommand{\oursshort}{ARDuP\xspace}

\newcommand{\more}[1]{{\scriptsize\textcolor[rgb]{0.22,0.71,0.29}{#1}}}

\newcommand{\name}{{ARDuP\xspace}}

\begin{document}

\maketitle
\thispagestyle{empty}
\pagestyle{empty}

\begin{abstract}
Sequential decision-making can be formulated as a text-conditioned video generation problem, where a video planner, guided by a text-defined goal, generates future frames visualizing planned actions, from which control actions are subsequently derived. 
In this work, we introduce Active Region Video Diffusion for Universal Policies (ARDuP), a novel framework for video-based policy learning that emphasizes the generation of active regions,~\textit{i.e.} potential interaction areas, enhancing the conditional policy's focus on interactive areas critical for task execution. This innovative framework integrates active region conditioning with latent diffusion models for video planning and employs latent representations for direct action decoding during inverse dynamic modeling. By utilizing motion cues in videos for automatic active region discovery, our method eliminates the need for manual annotations of active regions. We validate ARDuP's efficacy via extensive experiments on simulator CLIPort and the real-world dataset BridgeData v2, achieving notable improvements in success rates and generating convincingly realistic video plans.

\end{abstract}

\section{Introduction}
One critical objective of robotic learning is building a universal agent capable of performing a vast number of tasks across a diverse set of environments. Achieving this goal is challenging as the definition of a particular state or action may vary based on the task description. For instance, the state and action space of a robot tasked with navigating through a cluttered warehouse is differently defined than a robot whose purpose is to assemble intricate machinery. These variations demand a policy that not only provides a universal representation of the state space but also precisely identifies the actions necessary for any given task.

\begin{figure}[t]
  \centering

  \includegraphics[width=0.95\linewidth]{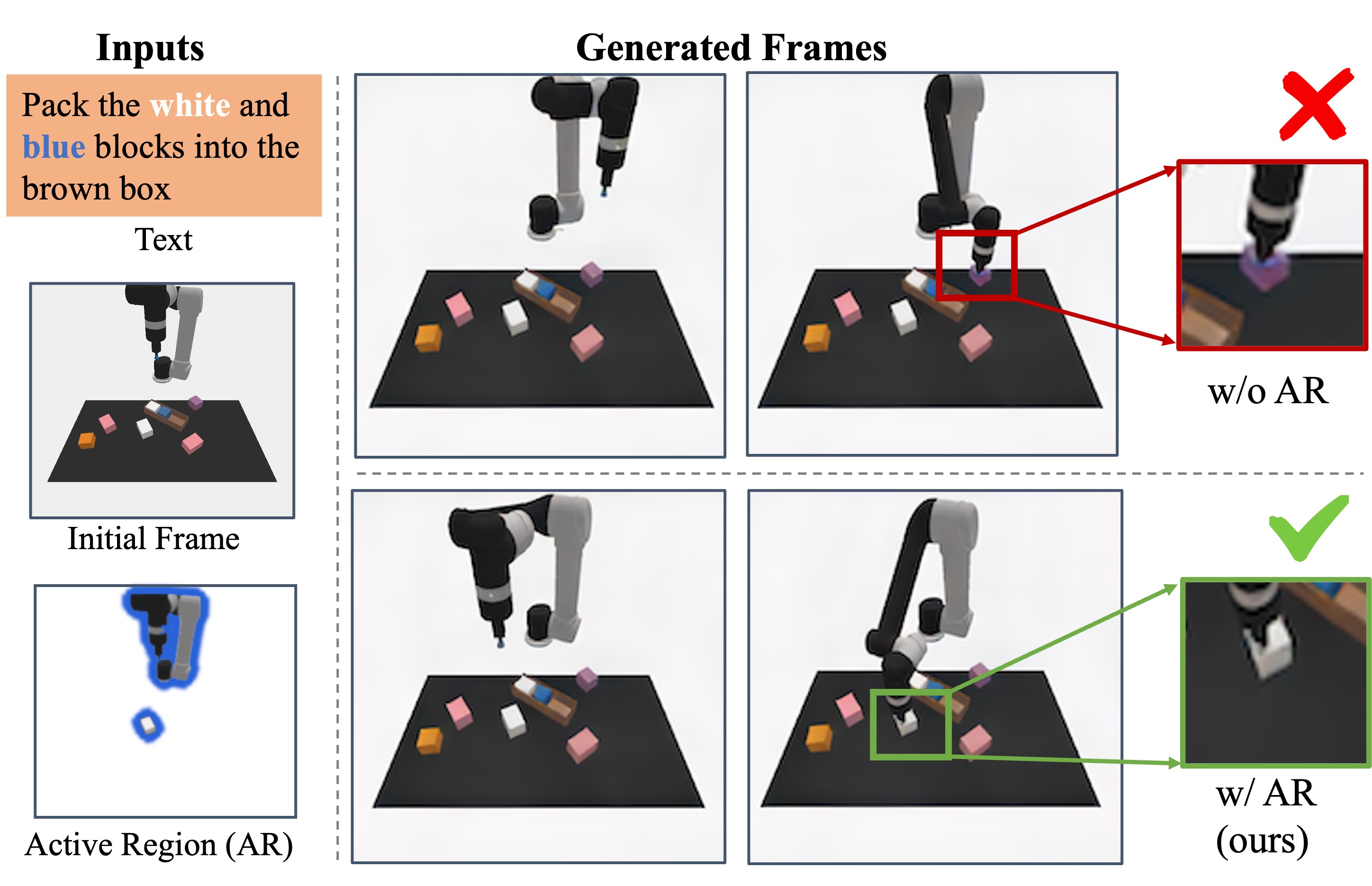}


 \caption{Given the task text and initial frame, we aim to generate a video as a planner. With active region conditioning (bottom left), our method ARDuP generates frames where the robot arm successfully picks up the white block (bottom), unlike the incorrect targeting of a purple block when w/o active region input (top), showing ARDuP's effectiveness in producing task-aligned video sequences.}
 
 \label{fig:teaser}
 \vspace{-0.2in}
\end{figure}

One promising direction is jointly using video and text descriptions to define a generalized state and action space. In these scenarios, a video generator is employed as a planner, 
which produces a sequential trajectory of frames as states given a text description of the immediate goal and an initial visual representation of the environment. Once the trajectory is generated, a policy conditioned on this trajectory (sequence of frames) is learned to infer the action taken between adjacent frames. The intuition is that using videos to represent the state space enables greater generalization across various tasks and environments. Recently, there has been considerable progress made in this field, with notable works~\cite{du2023video,du2024learning,ajay2024compositional} demonstrating success in tasks such as robot navigation and manipulation~\cite{zheng2024prise}. However, these methods often struggle to solve the task because they generate videos treating all pixels uniformly, often focusing on the wrong areas and neglecting to model pixels that are important for the policy. This can result in errors in generated frames, such as those seen on the top right of Figure~\ref{fig:teaser}. The robot picks up the wrong block (purple block) and can still reduce the loss in later frames by simply changing the color of the block to match the text description (e.g., changing purple to white).

\begin{figure*}[t]
  \centering
  \includegraphics[width=0.95\linewidth]{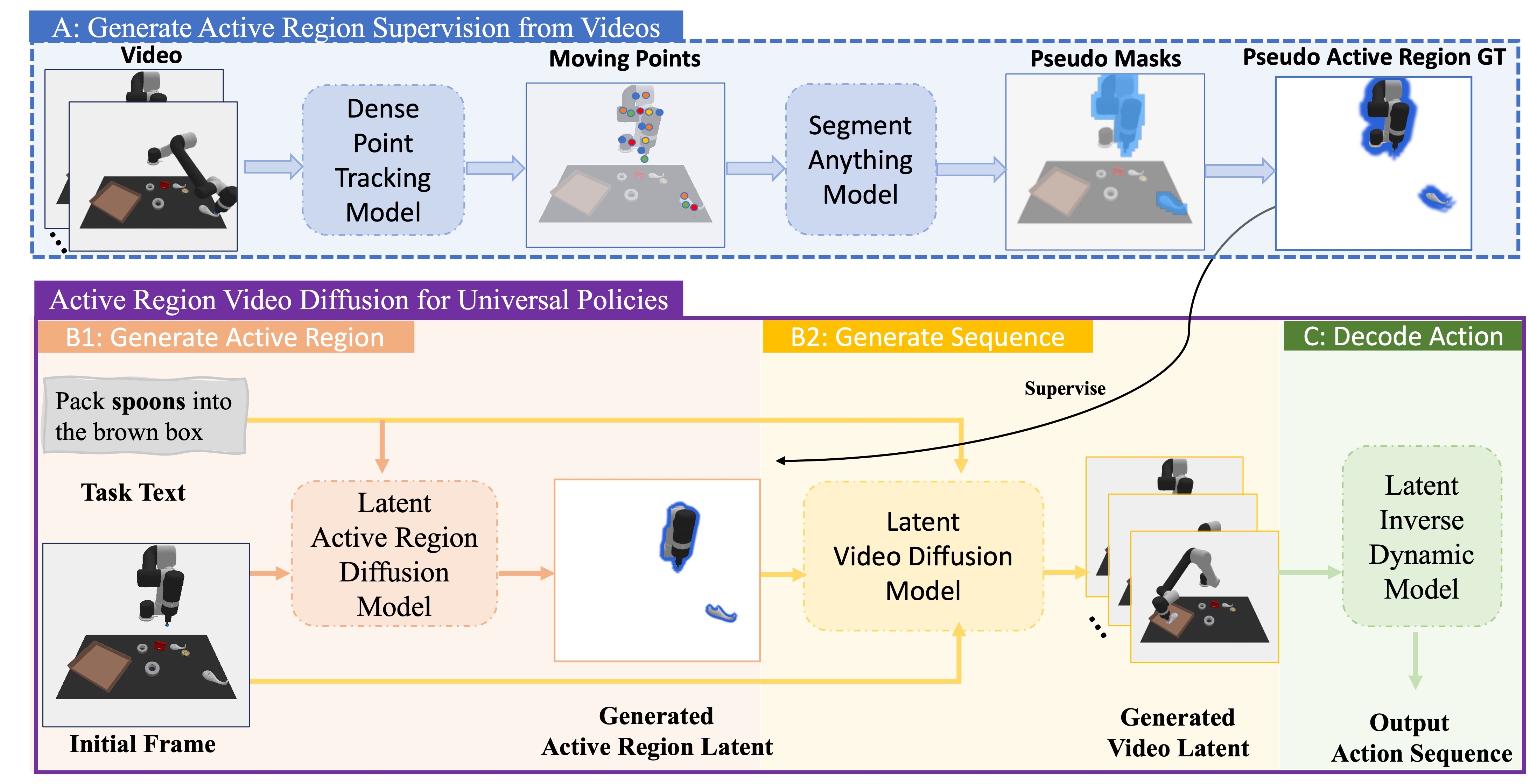}
  \caption{\textbf{Overview of our Active Region Video Diffusion for Universal Policies (ARDuP).} Starting with a video, we use Co-Tracker to identify moving points at the initial frame, which SAM then uses to generate pseudo masks of active regions. These pseudo masks delineate the pseudo active region, serving as supervision for training our Latent Active Region Diffusion Model. The generated latent active region informs the Latent Video Diffusion Model, resulting in a synthesized video latent sequence. Finally, the Latent Inverse Dynamic Model decodes the generated latent sequence into a corresponding action sequence.}
  
  \label{fig:vis_overview}
  \vspace{-0.2in}
\end{figure*}

To address this, our key insight is that only certain regions are important to solve the task; and these \textit{active regions}, which represent areas where objects are being interacted with, contain critical context that helps to counteract the errors seen in prior work. 
By emphasizing active regions during generation, we align the policy's attention to objects referred to by the given task description. As shown in Figure~\ref{fig:teaser}, our framework identifies the arm and the white box as active regions, which is highlighted in the image on the bottom left. This knowledge is then used to guide the video generation. As a result of this guidance, our approach successfully learns a policy to pick up the white block, while the baseline approach selects a purple block.

Formally, we propose Active Region Video Diffusion for Universal Policies (\oursshort), as shown in Figure~\ref{fig:vis_overview}. \oursshort consists of the following key components: 1) decomposed
video generation via (a) a latent active region diffusion model to generate the active region proposals, and (b) a latent 
video diffusion model which conditions on the given active region context to generate future trajectory latent codes; and 2) an inverse dynamics model to decode the synthesized trajectory latent codes into an action sequence. 

Since the ground truth active region is not available in the real-world scenario, we generate pseudo active regions from videos for training. Specifically, we employ Co-Tracker~\cite{karaev2023cotracker}, a pre-trained dense point tracking model, to identify moving trajectories. These moving points are fed to SAM~\cite{kirillov2023segment} to produce pseudo masks of active regions. This eliminates the need for manual annotations. In addition, we implement our video planner and inverse dynamics module in the latent space, which further reduces the computational cost.

We evaluate our method on public benchmarks, including the simulator CLIPort~\cite{shridhar2022cliport}, and the real-world dataset Bridgev2~\cite{walke2023bridgedata}. Specifically, we outperform the baseline by $21.3\%$, $17.2\%$, and $15.7\%$ for the Place Bowl, Pack Object, and Pack Pair tasks, respectively. We also conduct a detailed ablation study to illustrate the benefits of our proposed modules. The main contributions of this work are as follows:
\begin{itemize}
\item We propose a novel formulation for video-based policy learning that focuses on the generation of active regions, enhancing the conditional policy's attention to interactive areas critical for task execution. 
\item We introduce \oursshort, a latent framework that integrates active region conditioning with latent diffusion models for video planning, and adopt latent representations for direct action decoding via inverse dynamics modeling. By utilizing motion
cues in videos for automatic active region discovery, our method eliminates the need for manual annotations of active regions.
\item We validate \oursshort's efficacy via extensive experiments on CLIPort and real-world dataset BridgeData v2, achieving notable improvements in success rates and generating convincingly realistic video plans.
\end{itemize}

\section{Related Work}

\subsection{Video Diffusion Models for Decision Making}
Recent advancements in diffusion models have led to their application across various decision-making problems~\cite{ajay2022conditional,yang2024video}. For instance, Diffuser~\cite{janner2022planning} developed an unconditional diffusion model to generate trajectories comprising states and actions based on joints. As a pioneering approach, UniPi~\cite{du2024learning} casts the decision-making problem as a text-conditioned video generation problem and uses a video diffusion model as a universal policy. Extending UniPi, \cite{ko2023learning} leverages dense correspondence for action inference. And \cite{du2023video} incorporates VLMs with video diffusion models to facilitate the generation of long-horizon video plans. Existing works usually use the video diffusion model to generate every pixel indifferently of the videos. In comparison, we propose to emphasize the generation of task-relevant objects, which leads to policy more aligned with the task requirements.

\subsection{Active Region for Vision and Robotics}
 Active objects are objects being manipulated by the user and are important to the action being performed~\cite{pirsiavash2012detecting}. The idea that some regions are more important than others has been investigated in First Person Vision~\cite{damen2016you} and egocentric activity recognition~\cite{li2015delving,thakur2024leveraging}. In contrast, active regions for video-based policy are under-explored. We emphasize active regions in video generation and align the policy’s attention to active regions based on the task text. Since the ground truth active region is not available in
real world, we generate pseudo-active regions from videos via dense point tracking and SAM to eliminate manual labeling~\cite{karaev2023cotracker,huang2022learning,huang2019dynamic,he2023towards,huang2024point,huang2024uvis,kirillov2023segment}.

\section{Problem Formulation}
\subsection{Preliminaries}

Markov Decision Processes (MDPs)~\cite{puterman2014markov} play a crucial role in finding a solution for sequential decision-making problems. However, scaling algorithms that solve MDPs to diverse environments is challenging. This is because there is no standardized way to define states of the MDP as well as the difficulty defining an appropriate reward structure and the reliance of MDPs on environment-specific dynamics models.

The Unified Predictive Decision Process (UPDP) is introduced in~\cite{du2024learning} and aims to solve the issues that arise with an MDP. They do this by (1) using video as their state space, (2) utilizing recent advances~\cite{radford2021learning} in text-to-image understanding to use text to define the goal instead of an arbitrary reward and (3) developing a task-agnostic planning algorithm to find the action instead of relying on a predefined dynamics model. These three features enable UPDPs to scale across a wide variety of tasks in a way that MDPs cannot.

Formally, a UPDP is a tuple $\mathcal{G} = (\mathcal{X}, \mathcal{C}, H, \rho)$, where $\mathcal{X}$ is the observation space and each ${x_0, x_1, \dots, x_H} \in \mathcal{X}$ is an RGB frame, $\mathcal{C}$ is the set of task descriptions, $H$ is the task length and $\rho(\cdot|x_0, c)$ is a conditional video generator that synthesizes an $H$-step video $\{x_h\}^H_{h=1}\in\Delta(\mathcal{X}^H)$, where ${x_1, x_2, \dots, x_H}$ are predicted future frames conditioned on the first ground truth frame $x_0$ and the task description $c$. 

Given a UPDP $\mathcal{G}$, an action prediction algorithm is defined as $\mu(\cdot|\{x_h\}^H_{h=0}, c) \rightarrow \Delta(\mathcal{A}^H)$, where $\mathcal{A}^H$ represents an $H$-step action. This algorithm outputs an action sequence that aligns with the provided trajectory $\{x_h\}^H_{h=1}$ in the UPDP $\mathcal{G}$ for the task $c$. This algorithm is trained offline assuming access to a dataset of existing experiences $\mathcal{D} = \{(x_i, a_i)^{H-1}_{i=0}, x_H, c\}$. Given $\mathcal{D}$, $\rho(\cdot|x_0, c)$ and $\pi(\cdot|\{x_h\}^H_{h=0}, c)$ can be estimated.

\subsection{Latent Unified Predictive Decision Process conditioned on Active Region (LUPDP-AR)}
\label{sec:problem}
In the UPDP framework, the success of a task heavily relies on the action prediction algorithm, $\mu(\cdot|\{x_h\}^H_{h=0}, c)$. The action prediction, in turn, is conditioned on the images generated in the video generation stage. However, not all pixels generated in the video have an equal impact on the action. Active regions, which are typically objects that are the most likely to be interacted with, are more likely to have an impact on the action. By prioritizing focus on generation of the active regions we can better align the predicted actions with the task description $c$. We introduce an enhanced version of UPDP, which we call LUPDP-AR (Latent Unified Predictive Decision Process conditioned on Active Region). LUPDP-AR introduces active region conditioning to the video generator to foster a more interaction-aware policy.


Formally, an LUPDP-AR is defined as a tuple \(\hat{\mathcal{G}} = (\hat{\mathcal{X}}, \mathcal{C}, \hat{\mathcal{O}}, H, \phi)\), where \(\hat{\mathcal{X}}\) and \(\hat{\mathcal{O}}\) represent the latent spaces for RGB frames and active region frames, respectively. A frame encoder \(\mathcal{E}(\cdot)\) is adopted to map both RGB frames and active regions into these latent spaces~\cite{rombach2022high}. $\phi$ is a latent conditional generator that synthesizes an $H$-step latent trajectory $\{\hat{x}_h\}^H_{h=1}\in\Delta(\hat{\mathcal{X}}^H)$.

To ensure the accuracy of the active region in the generated trajectory we propose to condition the generator on the active region of the initial frame. Unlike the original UPDP, LUPDP-AR conditions on the latent representation of the active region $\hat{o} \in \hat{\mathcal{O}}$ as well as the latent of the initial frame $\hat{x}_0 \in \hat{\mathcal{X}}$ and the task description $c$, instead of conditioning on just the original frame $x_0$ and the task description $c$. We define this new video generator as \(\phi(\cdot|\hat{x}_0, c, \hat{o})\). Conditioning on active regions focuses the video generation process leading to more accurate actions for task completion.

To capture the active region in the initial frame, we define an active region generator as $\psi(\hat{o}|\hat{x}_0,c): \hat{\mathcal{X}} \times \mathcal{C} \rightarrow \hat{\mathcal{O}}$, which generates the latent of the active region $\hat{o}$ based on the latent of the first frame $\hat{x}_0$ and the task description $c$. Intuitively, our methodology decomposes the challenging trajectory generation by generating the active region of the initial frame first, followed by the generation of the full sequence under the guidance of the active region.

Given an LUPDP-AR $\hat{\mathcal{G}}$, we define a latent conditioned action prediction algorithm $\pi(\cdot|\{{\hat{x}_h}\}^H_{h=0}, c) \rightarrow \Delta(\mathcal{A}^H)$ where $\mathcal{A}^H$ is an $H$-step action sequence. $\pi$ only requires the latent of the generated images as input. This eliminates the need to generate RGB frames or decode a latent image, which is required in existing techniques such as~\cite{du2023video,du2024learning,ajay2024compositional}. We can still use a decoder, $\mathcal{R}$, to visually interpret the results, but it is not necessary for the action generation.

\section{Active Region Video Diffusion for Universal Policies (ARDuP)}
In this section, we will describe the details of our approach, Active Region Video Diffusion for Universal Policies (ARDuP), a tailored instantiation of the conditional diffusion models within our proposed LUPDP-AR framework. First, \name~finds the active region (Sec~\ref{subsec:pseudo}). Next, \name~utilizes a video generator conditioned on the active region as well as the language instruction and the initial frame to predict the future frames for the episode (Sec~\ref{subsec:diffusion}). Finally, given the generated video frames, \name~employs a latent inverse dynamic model to determine the actions the robot should take to complete the task (Sec~\ref{subsec:inv}).

\subsection{Active Region Generator}
\label{subsec:pseudo}
\subsubsection{Active Region Supervision from Videos}
As discussed in~\ref{sec:problem}, the key idea of our framework is to decompose the video generator $\rho(\cdot|x_0, c)$ into an active region generator $\psi(\hat{o}|\hat{x}_0,c)$ and a new video generator $\phi(\cdot|\hat{x}_0, c, \hat{o})$ conditioned on the active regions.
To train our active region generator without manual labeling, we propose to construct an active region dataset  \(\mathcal{D}_{\mathcal{O}}=\{\hat{x}_0,c,\hat{o}\}\) from video demonstrations by a large pre-trained dense point tracking model to pinpoint areas of significant activity throughout the episode.
By doing so, we can identify pseudo masks for active regions in each initial frame from \(\mathcal{D}\). 
The process of obtaining the active region from videos is shown in Figure~\ref{fig:vis_overview}.

Formally, given a video $\mathbf{V}=\{x_h\}^H_{h=0}$, we first obtain dense point trajectories $\mathcal{P}$ through a dense point tracking model $\mathcal{F}_p$. While any dense point tracking model can be used we adopt Co-Tracker~\cite{karaev2023cotracker} pretrained on real-world videos due to its efficiency~\cite{karaev2023cotracker}. Co-Tracker divides the initial frame into an $M\times M$ grid and tracks each point across $H+1$ frames. We denote the obtained dense point trajectories set as $\mathcal{P}  =\mathcal{F}_p (\mathbf{V})$, where $|\mathcal{P}| = M\times M$ is the size of all dense point trajectories. Each point trajectory consists of a single point's location from timestep 0 to $H$.

Given all point trajectories $\mathcal{P}$, we find the moving point trajectories by analyzing the change in pixel location. For each point trajectory $\mathbf{p} \in \mathcal{P}$, we compute the absolute movement $\Delta\mathbf{p}_h$ between two adjacent frames at timestep $h$ and $h-1$ as:
\begin{align}
\Delta\mathbf{p}_h = ||\mathbf{p}_h - \mathbf{p}_{h-1}||_2, \quad \text{for } h = 1, \ldots, H
\end{align}
where $\mathbf{p}_h$ and $\mathbf{p}_{h-1}$ are the point coordinates at time $h$ and $h-1$, respectively. The average movement $\Delta\bar{\mathbf{p}}$ of a point trajectory over $H$ timesteps is given by:
\begin{align}
\Delta\bar{\mathbf{p}} = \frac{1}{H} \sum_{h=1}^{H} \Delta\mathbf{p}_h
\end{align}

To identify moving point trajectories that exhibit significant
displacement, we apply a selection criterion based on $\tau$ an average movement threshold, denoted by $\tau$ as follows:
\begin{align}
\mathcal{P}_m = \{\mathbf{p} \in \mathcal{P} | \Delta\bar{\mathbf{p}} > \tau\}
\end{align}
where $\tau$ is a pre-selected threshold. $\mathcal{P}_m$ denotes the set of selected points whose average movement exceeds $\tau$. We then obtain coordinates of these points at the initial frame from $\mathcal{P}_m$ and denote this set as $\mathcal{P}_0$.

To generate the pseudo masks $\mathbf{M}$ for the active region at the initial frame, we feed the initial frame $x_0$ into the segmentation anything model (SAM)~\cite{kirillov2023segment} and prompt SAM with the coordinates at the initial frame $\mathcal{P}_0$. 

Given the derived pseudo mask $\mathbf{M}$ of the active regions, we define the pseudo active region frame $o$ at the initial frame $x_0$ as follows:
\begin{align}
    o = {x}_0\circ \mathbf{M} + x_b\circ(1-\mathbf{M})
\end{align}, where $\circ$ is element-wise multiplication and $x_b$ is a background frame with entirely white pixels. This process allows us to bypass the need for manual labeling.
Finally, we apply the pretrained image encoder $\mathcal{E}$ from Stable Diffusion~\cite{rombach2022high} to generate the latent representation of the active region $\hat{o}$.

\subsubsection{Latent Active Region Diffusion}
Once we have constructed our dataset $\mathcal{D}_{\mathcal{O}}$ we define a model to identify the active region at test time. \name~adopts a conditioned latent diffusion model $\psi(\hat{o}|\hat{x}_o, c;\theta_{\psi})$ to accomplish this. 
The diffusion model takes as input both a textual description of a task and the latent representation of the initial frame and outputs the active region. 
It is necessary to use both inputs because only together can the model accurately pinpoint areas of potential interaction. Without the frame, the model would not know where items are positioned and without the text description, the model would not understand the context of the task.

Diffusion models offer flexibility, allowing the selection of various active regions, which is crucial for tasks with multiple valid solutions. For example, in a task described as ``put blue blocks into the box'', any unplaced blue block can be identified as a valid active region. This flexibility is particularly beneficial in dynamic environments where multiple paths to task completion may exist.

During inference, we generate the active region using the trained active region generator $\psi(\hat{o}|\hat{x}_o, c;\theta_{\psi})$. The generated active region, along with the provided text and initial frame, is then input into the video planner $\phi(\cdot|\hat{x}_0, c, \hat{o};\theta_{\phi})$ for generating the latent sequence, as detailed below.

\begin{figure*}[t]
  \centering
  \includegraphics[width=0.95\linewidth]{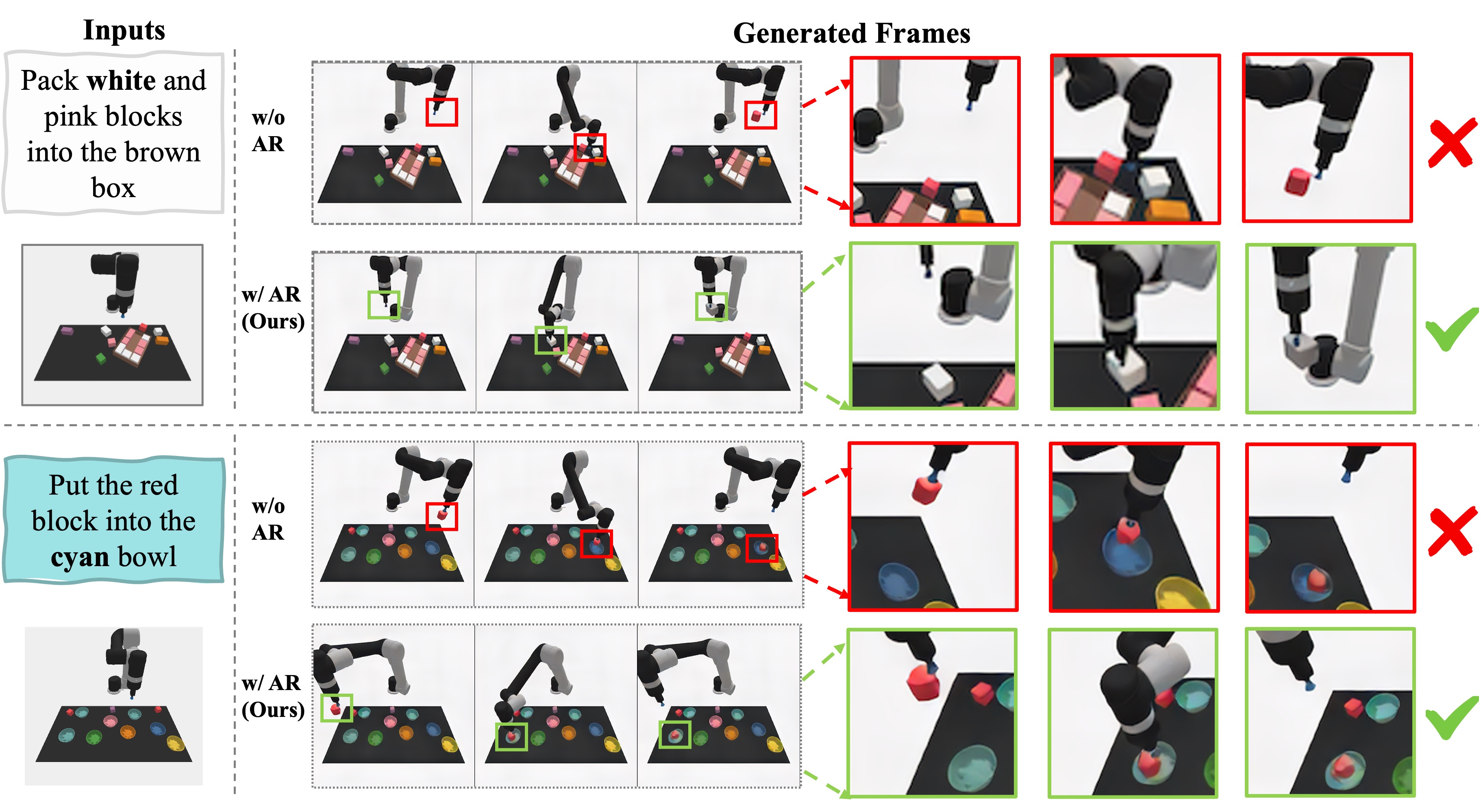}
  \caption{\textbf{Qualitative comparison of the generated video plans on CLIPort~\cite{shridhar2022cliport} unseen test tasks.} Our
method successfully generates a video that packs the white
block, while the counterpart w/o active region generates frames that pick the
wrong object (the red block) or fails to correctly generate
the arm position for grasping objects. We demonstrate higher visual
quality, especially around objects of interest, thanks to the
guidance provided by our active regions. }
  \label{fig:vis_cliport}
  \vspace{-0.2in}
\end{figure*}

\subsection{Universal Video Planner conditioned on Active Region}
\label{subsec:diffusion}
Motivated by the advances of text-to-video models~\cite{ho2022imagen}, we develop a latent video diffusion model to serve as our trajectory planner. 
This module is designed to accurately generate future video latent representations based on an initial frame and a textual description of the task at hand as well as the active region. 

Different from previous works, our proposed video generation model, $\phi(\cdot|\hat{x}_0, c, \hat{o};\theta_{\phi})$, integrates the active region alongside the initial frame and text description for conditioning. 
The key insight is that by guiding our model with the active region, it can focus its generation efforts on areas crucial to the described task, enhancing the relevance and precision of the generated sequence.

To perform this integration, we concatenate each frame's latent with the latent of the active region $\hat{o}$. We do this in addition to the concatenation of the active region latent with the latent of the initial frame $\hat{x}_0$. This dual concatenation strategy ensures that, the model's denoising process for each frame aligns not only with the initial observation but also with the active region latents. 

During inference, the input active region of $\phi(\cdot|\hat{x}_0, c, \hat{o};\theta_{\phi})$ is generated from the trained active region generator $\psi(\hat{o}|\hat{x}_o, c;\theta_{\psi})$ given input text and initial frame. During training of $\phi(\cdot|\hat{x}_0, c, \hat{o};\theta_{\phi})$, the input active region is obtained from the training videos as detailed above.

\subsection{Task Specific Action Decoding from Latent Sequence}

\label{subsec:inv}
Finally, we train a compact, task-specific latent inverse dynamics model to convert synthesized latent sequences directly into action sequences.

\subsubsection{Latent Inverse Dynamics Model} 
Given two adjacent generated frame latents $\hat{x}_h$ and $\hat{x}_{h+1}$, our latent inverse dynamics module predicts the action $a_h$. The training of the inverse dynamics is independent of the video planner and can be done on a separate, smaller, and potentially suboptimal dataset generated by a simulator. Both the video planner and the inverse dynamics model utilize the same encoder $\mathcal{E}$ to map the RGB frame into the latent space. This ensures a consistent latent embedding space for both the video generation and the action decoding phases.

During inference, we decode actions directly from latent frames. Our latent-based inverse dynamic design represents a significant advancement over the existing RGB-based inverse dynamic models~\cite{du2024learning,du2023video,ajay2024compositional}, which rely on RGB frames for action decoding. A main challenge with training video diffusion models to output RGB frames directly is the computational cost. Our latent dynamics module, on the other hand, bypasses the need to generate RGB frames because it can compute the action directly from the latent. This strategy not only streamlines the decoding process but also makes \name's architecture more efficient and compact. For the purpose of visual interpretability, we can still employ the decoder $\mathcal{R}$ to transform these synthesized latents back into RGB frames if necessary.

\subsubsection{Action Execution} During the execution phase of \name, we start with $x_0$ and $c$. Next we generate the active region latent 
$\hat{o}$, followed by the synthesis of $H$-step latent sequence. The generated latents are then fed into our trained latent inverse dynamics model to produce the corresponding $H$ actions. To enhance computational efficiency, we employ an open-loop control strategy by sequentially performing actions from the initially predicted action sequence.

\section{Experiments}

\begin{figure*}[t]
  \centering
  \includegraphics[width=0.90\linewidth]{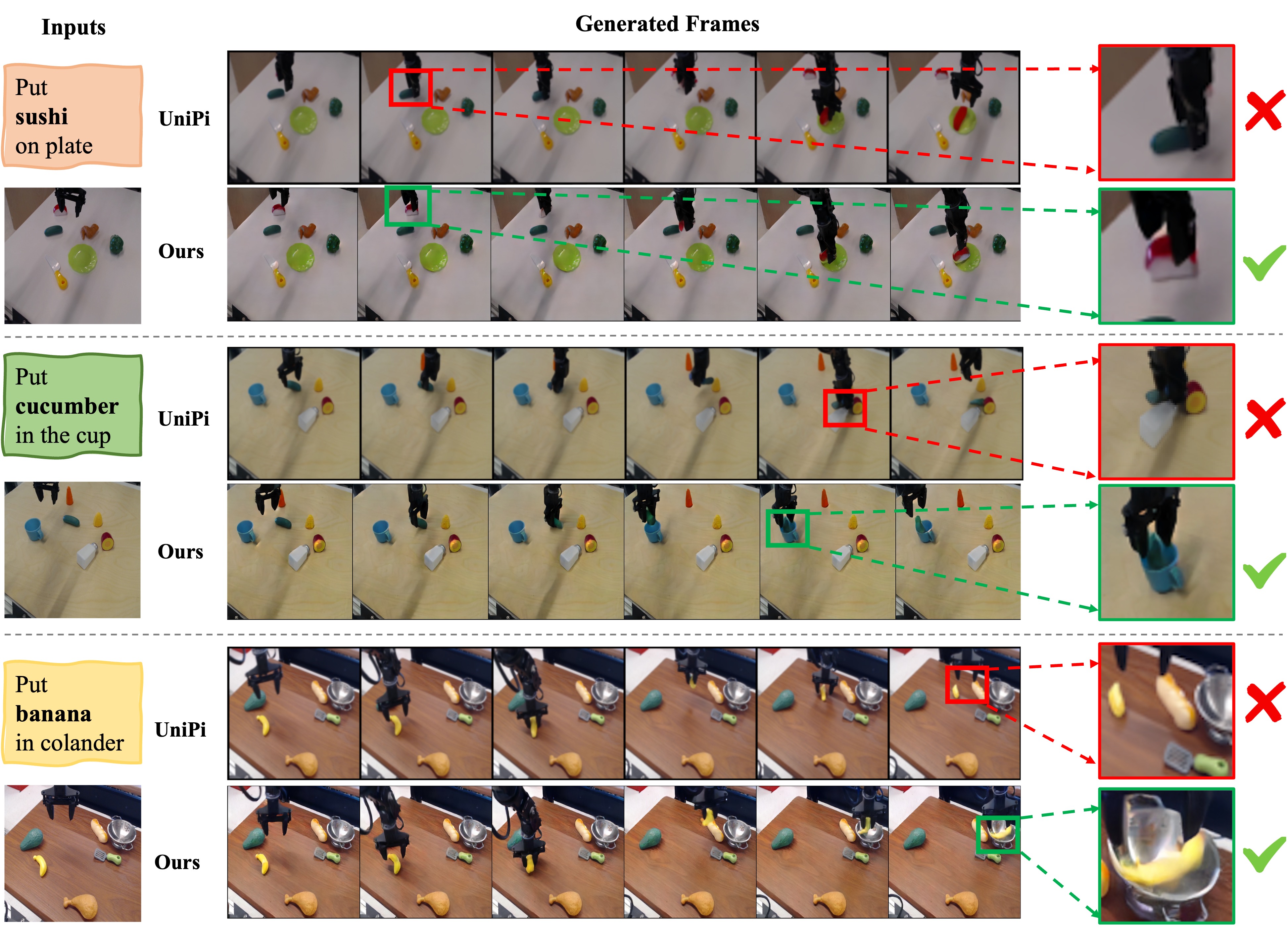}
  \caption{\textbf{Qualitative comparison of the generated video plans on BridgeData v2~\cite{walke2023bridgedata}.} Our method successfully picks up the correct object
(sushi) or places the object in the appropriate location (in
the cup or colander), while the UniPi* baseline selects
the wrong object (cucumber) or places it incorrectly (on the
desk), showing the advantage of our method.}
  \label{fig:vis_bridge}
  \vspace{-0.2in}
\end{figure*}

In the following sections, we evaluate our method on two public robot manipulation datasets, CLIPort~\cite{shridhar2022cliport} and BridgeData v2~\cite{walke2023bridgedata}. First, we describe the implementation details in Sec.~\ref{sec:imple}. In Sec.~\ref{sec:cliport} we evaluate the generalization ability of our method across multiple tasks. We provide these results on CLIPort. Sec.~\ref{sec:bridge} demonstrates the transfer ability of our method to real-world complex scenarios. We report results on BridgeData v2. Finally, ablation studies are provided in Sec.~\ref{sec:ablation}, and limitations are discussed in Sec.~\ref{sec:limit}.

\subsection{Implementation Details}
\label{sec:imple}
For the active region generation, we use Co-Tracker~\cite{karaev2023cotracker}, which requires a grid size to determine the total number of points to track. \name~uses a grid size of $M=60$. We also set a threshold $\tau=2$ for the minimum movement of a point trajectory to be considered active. We conduct data preprocessing of connected component labeling and analysis to retain pseudo masks for manipulated objects while excluding those of the robot arm. The encoder $\mathcal{E}$ and decoder $\mathcal{R}$ are trained using a variational autoencoder as described in Stable Diffusion~\cite{rombach2022high}. Images are resized to 512$\times$512 when fed into the encoder, and the encoder produces a latent representation of shape $128\times128\times4$. The diffusion models and the inverse dynamic model are trained from scratch on 8 A100 GPUs.

To find the active region, we utilize the T5-XXL model~\cite{raffel2020exploring} and a modified UNet architecture from~\cite{dhariwal2021diffusion}. We use the same architecture for the latent video diffusion model but utilize the temporal modules modification proposed in~\cite{ko2023learning} for generating $H=6$ frame latents. We adopt a learning rate of $3 \times 10^{-4}$ and a batch size of 96 for training.

Our latent inverse dynamic model consists of convolutional layers with skip connections followed by a linear layer to project features into 7-dimensional action vectors for manipulation tasks. This latent action decoder is trained with a learning rate of $5 \times 10^{-4}$ and a batch size of 3072.

\begin{table}[t]
\centering
\caption{Success rate in completing tasks across multiple environments in CLIPort~\cite{shridhar2022cliport}. When trained across various multi-task environments, our approach shows strong generalization capabilities to unseen environments, significantly outperforming the UniPi baseline~\cite{du2024learning}. UniPi* is reproduced based on the reference code provided by the authors. }
\resizebox{0.48\textwidth}{!}{
\begin{tabular}{@{}llll@{}}
\toprule
Model                   & Place Bowl          & Pack Object        & Pack Pair  \\ \midrule
State + Transformer BC~\cite{brohan2022rt}  & 9.8  & 21.7 & 1.3 \\
Image + Transformer BC~\cite{brohan2022rt}  & 5.3 & 5.7  & 7.8 \\
Image + TT~\cite{janner2021offline}              & 4.9 & 19.8 & 2.3 \\
Diffuser~\cite{janner2022planning}                & 14.8& 15.9 & 10.5 \\ \midrule
UniPi*~\cite{du2024learning}           & 65.4  & 51.8 & 30.9 \\ 
Ours            & \textbf{86.7}\more{(+21.3)}  & \textbf{69.0}\more{(+17.2)}  & \textbf{46.6}\more{(+15.7)}\\ \bottomrule
\end{tabular}
}
\label{tab:performance_comparison}
\vspace{-0.15in}
\end{table}

\subsection{Multi-Environment Transfer on CLIPort}
\label{sec:cliport}

\textit{Setup.} CLIPort~\cite{shridhar2022cliport} is an extended benchmark of language-grounded tasks for manipulation in the simulator Ravens~\cite{zeng2021transporter}. We train our models on 110k demonstrations spanning 11 distinct tasks and evaluate on 3 unseen tasks in CLIPort. Testing on 3 unseen tasks allows us to evaluate the model's ability to generalize to new tasks. We follow the multi-environment transfer setup in UniPi~\cite{du2024learning}. We report the final task completion success rate across new instances of the test environment and associated language prompts.

\textit{Baseline.} UniPi~\cite{du2024learning} generates RGB video frames and employs an RGB-based inverse dynamic model. We provide results on our reproduction of UniPi based on the reference code provided by the authors, denoted as UniPi*. 
We aim to match the number of in-domain training demonstrations. Nevertheless, the performance discrepancy could still be the result of differences in training data, as the training and pre-training data for UniPi is not publicly available.
Additionally, we also compare with the same set of imitation learning baselines listed in~\cite{du2024learning}.  
We refer readers to the original paper for a detailed description of the baselines.

\textit{Results.} In Table~\ref{tab:performance_comparison}, we show the performance of our method compared with established baselines on three unseen tasks in CLIPort. Notably, our method significantly improves the success rate over the UniPi baseline by $21.3\%$, $17.2\%$, and $15.7\%$ for the Place Bowl, Pack Object, and Pack Pair tasks, resp. This performance is driven by better generation due to our active region-conditioned diffusion models and improved execution produced by the latent inverse dynamics module. We visualize generated video plans in Figure~\ref{fig:vis_cliport}, highlighting the visual improvements of \name~over a counterpart model without active region conditioning. The counterpart is the same latent video diffusion model as ours except for conditioning on the active region. Using an active region results in higher visual quality, especially around objects of interest. For example, our method successfully generates a video that packs the white block, while the counterpart generates frames that either pick the wrong object or fail to generate the arm position correctly.

\begin{table}[b!]
\centering
\caption{Effect of Active Region Quality on Performance. The higher-quality active region during training and testing phases, the more improvement in task success rates, highlighting the significance of the active region.}

\label{tab:mask}
\renewcommand{\tabcolsep}{4pt}
\resizebox{0.48\textwidth}{!}{
\begin{tabular}{@{}lllccc@{}}
\toprule
AR (Trn)  &AR (Test)     & Video Planner Setting &Place Bowl   & Pack Object  & Pack Pair   \\ \midrule
-&  -         &w/o Active Region    & 83.4  & 67.7 & 38.0 \\ \midrule
Unsupervised &Predicted&+Active Region & 86.7  & 69.0 & 46.6 \\ 
& &$\Delta$    & +3.3  & +1.3 & +8.6 \\ \midrule
Supervised  &Predicted  &+Active Region   & 93.3  & 79.6 & 51.7 \\ 
 &  &$\Delta$  & +9.9  & +11.9 &  +13.7\\ \midrule
- &GT  &+Active Region   & 100.0  & 92.1 & 62.8 \\ 
&    &$\Delta$  & +16.6  & +24.4 &  +24.8\\ \bottomrule
\end{tabular}
}
\end{table}

\subsection{Real-world Transfer on BridgeData v2}
\label{sec:bridge}
\textit{Setup.} BridgeData v2~\cite{walke2023bridgedata} is a large and diverse real-world dataset for robotic manipulation. Bridge's examples show a wide variety of skills using various unique objects in diverse environments. This dataset includes 60,096 trajectories with natural language instructions, gathered from 24 environments with a widely accessible, cost-effective robot. We use 95\% of the data for training and the rest for evaluation.

\textit{Results.} As illustrated in Figure~\ref{fig:vis_bridge}, our method is able to generate realistic video plans that effectively handle chaotic environments, diverse language instructions, and varying camera poses. In the examples shown, our approach is continuously able to pick up the correct object and place it in the correct location. In contrast, the UniPi baseline often selects the wrong object and places it incorrectly. For example in the top example instead of picking up the sushi the UniPi baseline picks up a cucumber. In the bottom example, our method successfully finds the colander while the UniPi baseline places the banana back on the table. These results showcase the robust adaptability of our method across a wide range of complex real-world scenarios, highlighting its superior performance in understanding tasks in challenging settings. 

\begin{figure}[t]
  \centering

  \includegraphics[width=1.0\linewidth]{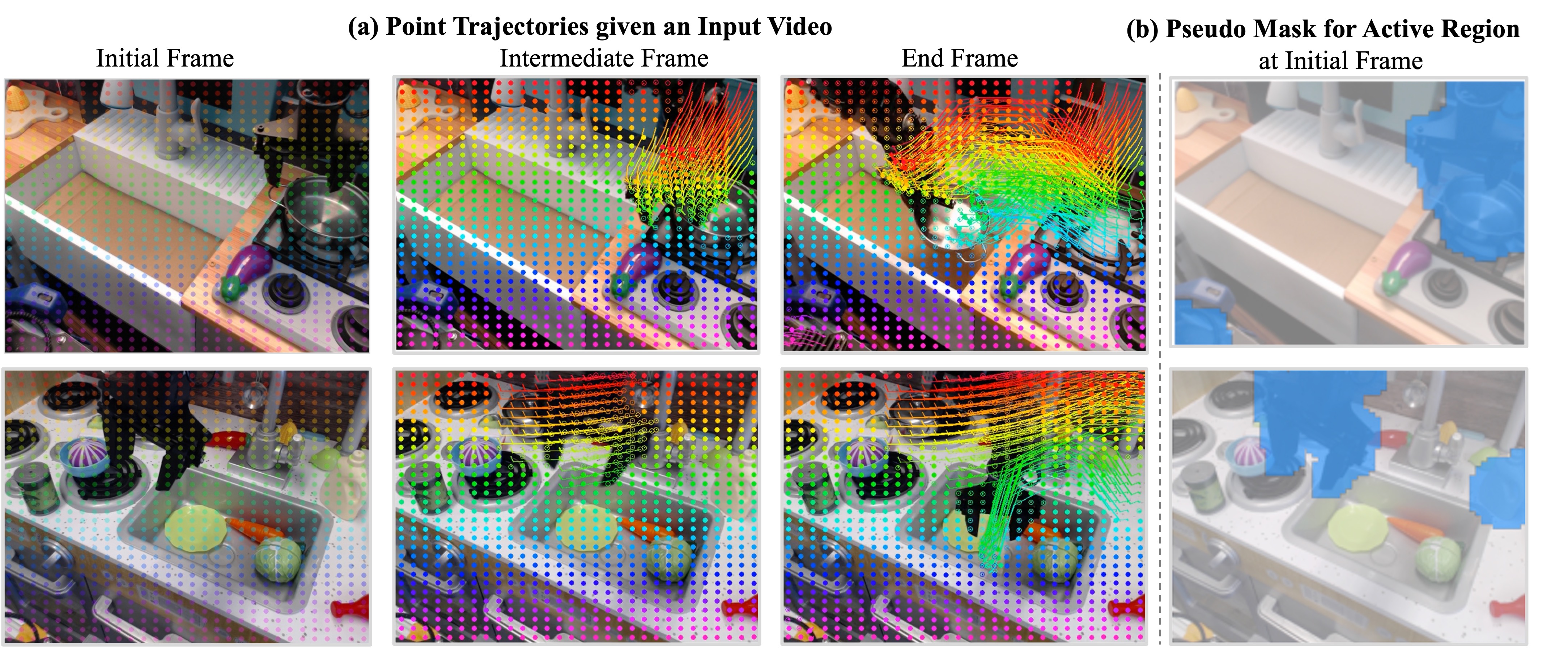}

 \caption{\textbf{Visualization of pseudo active region in BridgeData v2.} We visualize the dense point trajectories and the obtained pseudo masks of the active region in BridgeData v2~\cite{walke2023bridgedata}.}
 \label{fig:vis_pseudo_bridge}
 \vspace{-0.2in}
\end{figure}

Figure~\ref{fig:vis_pseudo_bridge} showcases the high-quality pseudo masks generated by our method in real video scenarios. This is achieved without the need for manual annotations. This visualization underscores the effectiveness of our approach in accurately identifying moving objects directly from video data.

\begin{figure}[t]
  \centering

  \includegraphics[width=0.9\linewidth]{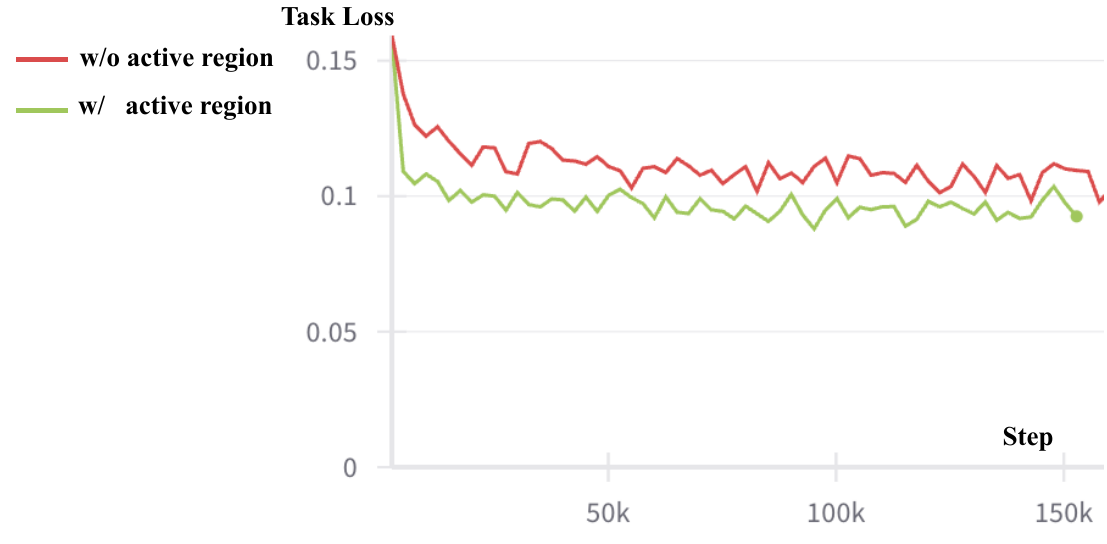}
\vspace{-0.1in}
 \caption{\textbf{Task Loss on CLIPort~\cite{shridhar2022cliport} test set.} Our method (green) shows significantly lower task loss compared to that without the active region (red), underscoring the active region’s role in enhancing generation quality.}
 \label{fig:vis_ablation_active}
 \vspace{-0.1in}
\end{figure}

\subsection{Ablation Study}
\label{sec:ablation}

\subsubsection{Quantitative evaluation of generation quality for control tasks} To quantitatively evaluate the video generation
quality for control-specific tasks, we propose a generation
metric named task loss. Our task loss is calculated by feeding
the generated video latents into our pre-trained latent inverse
dynamic module for action decoding and then computing the
L1 error between the decoded actions and the ground truth
actions. Task loss provides a metric for how well the generated video is aligned with the expected action sequence. A visual comparison of the task loss curves for the CLIPort test set is shown in Figure~\ref{fig:vis_ablation_active}. The green curve which represents our model when conditioned on active regions has a significantly lower task loss than the red line, which is the counterpart with no active region input. This result emphasizes the active region’s role in enhancing generation quality and enabling more precise action decoding.

\subsubsection{Effect of Active Region Quality on Performance} Table~\ref{tab:mask} shows the impact of active region quality on the success rate. Incorporating active regions in the unsupervised way described in the paper yielded a performance boost of +3.3\%, +1.3\%, and +8.6\%. This boost is even more (+9.9\%, +11.9\%, and +13.7\%, respectively) when the active region prediction model is trained with ground truth active regions. Furthermore, using ground truth active regions during inference led to even more substantial boosts (+16.6\%, +24.4\%, and +24.8\%, respectively). These results show that the task performance continues to improve with higher quality active regions and underscore its critical role in enhancing task success rates through video-based policy learning.

\subsection{Limitations and Future work}
\label{sec:limit}
One limitation of our approach is that pseudo-active region quality depends on Co-Tracker, which may struggle with long-horizon tasks or complex backgrounds. Advanced dense point tracking models could improve pseudo label quality, and noisy label learning techniques might stabilize training. Alternatively, text-guided segmentation could be used, where the vision-language model identifies the target object based on the task, followed by segmentation using tools like Segment Anything. We leave these as future work.

\section{Conclusion}
In this work, we present ARDuP, a novel framework for video-based policy learning by prioritizing active region generation. The video planner in ARDuP is a decomposed video generator: firstly, active regions are generated via a latent conditional diffusion model based on text input and an initial frame; secondly, video latents are synthesized with additional active region conditioning. The synthesized latent sequences are then decoded into action sequences through a latent inverse dynamic model. During training, we obtain active region supervision without human annotation by using Co-Tracker for moving point tracking and SAM for pseudo mask generation. ARDuP is validated on the CLIPort simulator and the real-world BridgeData v2 dataset, showing notable success rate improvement and the strong ability to generate realistic and action-centric video plans.

\smallskip
\noindent\textbf{Acknowledgements}
This work was partially supported by NSF CAREER Award (\#2238769) to AS.

{\small
\bibliographystyle{IEEEtran}
\bibliography{vid_robot_2024}

\begin{thebibliography}{10}
\providecommand{\url}[1]{#1}
\csname url@rmstyle\endcsname
\providecommand{\newblock}{\relax}
\providecommand{\bibinfo}[2]{#2}
\providecommand\BIBentrySTDinterwordspacing{\spaceskip=0pt\relax}
\providecommand\BIBentryALTinterwordstretchfactor{4}
\providecommand\BIBentryALTinterwordspacing{\spaceskip=\fontdimen2\font plus
\BIBentryALTinterwordstretchfactor\fontdimen3\font minus \fontdimen4\font\relax}
\providecommand\BIBforeignlanguage[2]{{%
\expandafter\ifx\csname l@#1\endcsname\relax
\typeout{** WARNING: IEEEtran.bst: No hyphenation pattern has been}%
\typeout{** loaded for the language `#1'. Using the pattern for}%
\typeout{** the default language instead.}%
\else
\language=\csname l@#1\endcsname
\fi
#2}}

\bibitem{du2023video}
Y.~Du, S.~Yang, P.~Florence, F.~Xia, A.~Wahid, brian ichter, P.~Sermanet, T.~Yu, P.~Abbeel, J.~B. Tenenbaum, L.~P. Kaelbling, A.~Zeng, and J.~Tompson, ``Video language planning,'' in \emph{The Twelfth International Conference on Learning Representations}, 2024.

\bibitem{du2024learning}
Y.~Du, S.~Yang, B.~Dai, H.~Dai, O.~Nachum, J.~Tenenbaum, D.~Schuurmans, and P.~Abbeel, ``Learning universal policies via text-guided video generation,'' \emph{Advances in Neural Information Processing Systems}, vol.~36, 2023.

\bibitem{ajay2024compositional}
A.~Ajay, S.~Han, Y.~Du, S.~Li, A.~Gupta, T.~Jaakkola, J.~Tenenbaum, L.~Kaelbling, A.~Srivastava, and P.~Agrawal, ``Compositional foundation models for hierarchical planning,'' \emph{Advances in Neural Information Processing Systems}, vol.~36, 2024.

\bibitem{zheng2024prise}
R.~Zheng, C.-A. Cheng, H.~D. III, F.~Huang, and A.~Kolobov, ``{PRISE}: {LLM}-style sequence compression for learning temporal action abstractions in control,'' in \emph{Forty-first International Conference on Machine Learning}, 2024.

\bibitem{karaev2023cotracker}
N.~Karaev, I.~Rocco, B.~Graham, N.~Neverova, A.~Vedaldi, and C.~Rupprecht, ``Cotracker: It is better to track together,'' \emph{arXiv preprint arXiv:2307.07635}, 2023.

\bibitem{kirillov2023segment}
A.~Kirillov, E.~Mintun, N.~Ravi, H.~Mao, C.~Rolland, L.~Gustafson, T.~Xiao, S.~Whitehead, A.~C. Berg, W.-Y. Lo, \emph{et~al.}, ``Segment anything,'' \emph{arXiv preprint arXiv:2304.02643}, 2023.

\bibitem{shridhar2022cliport}
M.~Shridhar, L.~Manuelli, and D.~Fox, ``Cliport: What and where pathways for robotic manipulation,'' in \emph{Conference on Robot Learning}.\hskip 1em plus 0.5em minus 0.4em\relax PMLR, 2022, pp. 894--906.

\bibitem{walke2023bridgedata}
H.~R. Walke, K.~Black, T.~Z. Zhao, Q.~Vuong, C.~Zheng, P.~Hansen-Estruch, A.~W. He, V.~Myers, M.~J. Kim, M.~Du, \emph{et~al.}, ``Bridgedata v2: A dataset for robot learning at scale,'' in \emph{Conference on Robot Learning}.\hskip 1em plus 0.5em minus 0.4em\relax PMLR, 2023, pp. 1723--1736.

\bibitem{ajay2022conditional}
A.~Ajay, Y.~Du, A.~Gupta, J.~Tenenbaum, T.~Jaakkola, and P.~Agrawal, ``Is conditional generative modeling all you need for decision-making?'' \emph{arXiv preprint arXiv:2211.15657}, 2022.

\bibitem{yang2024video}
S.~Yang, J.~Walker, J.~Parker-Holder, Y.~Du, J.~Bruce, A.~Barreto, P.~Abbeel, and D.~Schuurmans, ``Video as the new language for real-world decision making,'' \emph{arXiv preprint arXiv:2402.17139}, 2024.

\bibitem{janner2022planning}
M.~Janner, Y.~Du, J.~B. Tenenbaum, and S.~Levine, ``Planning with diffusion for flexible behavior synthesis,'' \emph{arXiv preprint arXiv:2205.09991}, 2022.

\bibitem{ko2023learning}
P.-C. Ko, J.~Mao, Y.~Du, S.-H. Sun, and J.~B. Tenenbaum, ``Learning to act from actionless videos through dense correspondences,'' \emph{arXiv preprint arXiv:2310.08576}, 2023.

\bibitem{pirsiavash2012detecting}
H.~Pirsiavash and D.~Ramanan, ``Detecting activities of daily living in first-person camera views,'' in \emph{2012 IEEE conference on computer vision and pattern recognition}.\hskip 1em plus 0.5em minus 0.4em\relax IEEE, 2012, pp. 2847--2854.

\bibitem{damen2016you}
D.~Damen, T.~Leelasawassuk, and W.~Mayol-Cuevas, ``You-do, i-learn: Egocentric unsupervised discovery of objects and their modes of interaction towards video-based guidance,'' \emph{Computer Vision and Image Understanding}, vol. 149, pp. 98--112, 2016.

\bibitem{li2015delving}
Y.~Li, Z.~Ye, and J.~M. Rehg, ``Delving into egocentric actions,'' in \emph{Proceedings of the IEEE conference on computer vision and pattern recognition}, 2015, pp. 287--295.

\bibitem{thakur2024leveraging}
S.~Thakur, C.~Beyan, P.~Morerio, V.~Murino, and A.~Del~Bue, ``Leveraging next-active objects for context-aware anticipation in egocentric videos,'' in \emph{Proceedings of the IEEE/CVF Winter Conference on Applications of Computer Vision}, 2024, pp. 8657--8666.

\bibitem{huang2022learning}
S.~Huang, L.~Yang, B.~He, S.~Zhang, X.~He, and A.~Shrivastava, ``Learning semantic correspondence with sparse annotations,'' in \emph{European Conference on Computer Vision}.\hskip 1em plus 0.5em minus 0.4em\relax Springer, 2022, pp. 267--284.

\bibitem{huang2019dynamic}
S.~Huang, Q.~Wang, S.~Zhang, S.~Yan, and X.~He, ``Dynamic context correspondence network for semantic alignment,'' in \emph{Proceedings of the IEEE/CVF International Conference on Computer Vision}, 2019, pp. 2010--2019.

\bibitem{he2023towards}
B.~He, X.~Yang, H.~Wang, Z.~Wu, H.~Chen, S.~Huang, Y.~Ren, S.-N. Lim, and A.~Shrivastava, ``Towards scalable neural representation for diverse videos,'' in \emph{Proceedings of the IEEE/CVF Conference on Computer Vision and Pattern Recognition}, 2023, pp. 6132--6142.

\bibitem{huang2024point}
S.~Huang, D.-A. Huang, Z.~Yu, S.~Lan, S.~Radhakrishnan, J.~M. Alvarez, A.~Shrivastava, and A.~Anandkumar, ``What is point supervision worth in video instance segmentation?'' in \emph{Proceedings of the IEEE/CVF Conference on Computer Vision and Pattern Recognition Workshops (CVPRW)}, 2024, pp. 2671--2681.

\bibitem{huang2024uvis}
S.~Huang, S.~Suri, K.~Gupta, S.~S. Rambhatla, S.-n. Lim, and A.~Shrivastava, ``Uvis: Unsupervised video instance segmentation,'' in \emph{Proceedings of the IEEE/CVF Conference on Computer Vision and Pattern Recognition Workshops (CVPRW)}, 2024, pp. 2682--2692.

\bibitem{puterman2014markov}
M.~L. Puterman, \emph{Markov decision processes: discrete stochastic dynamic programming}.\hskip 1em plus 0.5em minus 0.4em\relax John Wiley \& Sons, 2014.

\bibitem{radford2021learning}
A.~Radford, J.~W. Kim, C.~Hallacy, A.~Ramesh, G.~Goh, S.~Agarwal, G.~Sastry, A.~Askell, P.~Mishkin, J.~Clark, \emph{et~al.}, ``Learning transferable visual models from natural language supervision,'' in \emph{International conference on machine learning}.\hskip 1em plus 0.5em minus 0.4em\relax PMLR, 2021, pp. 8748--8763.

\bibitem{rombach2022high}
R.~Rombach, A.~Blattmann, D.~Lorenz, P.~Esser, and B.~Ommer, ``High-resolution image synthesis with latent diffusion models,'' in \emph{Proceedings of the IEEE/CVF conference on computer vision and pattern recognition}, 2022, pp. 10\,684--10\,695.

\bibitem{ho2022imagen}
J.~Ho, W.~Chan, C.~Saharia, J.~Whang, R.~Gao, A.~Gritsenko, D.~P. Kingma, B.~Poole, M.~Norouzi, D.~J. Fleet, \emph{et~al.}, ``Imagen video: High definition video generation with diffusion models,'' \emph{arXiv preprint arXiv:2210.02303}, 2022.

\bibitem{raffel2020exploring}
C.~Raffel, N.~Shazeer, A.~Roberts, K.~Lee, S.~Narang, M.~Matena, Y.~Zhou, W.~Li, and P.~J. Liu, ``Exploring the limits of transfer learning with a unified text-to-text transformer,'' \emph{The Journal of Machine Learning Research}, vol.~21, no.~1, pp. 5485--5551, 2020.

\bibitem{dhariwal2021diffusion}
P.~Dhariwal and A.~Nichol, ``Diffusion models beat gans on image synthesis,'' \emph{Advances in neural information processing systems}, vol.~34, pp. 8780--8794, 2021.

\bibitem{brohan2022rt}
A.~Brohan, N.~Brown, J.~Carbajal, Y.~Chebotar, J.~Dabis, C.~Finn, K.~Gopalakrishnan, K.~Hausman, A.~Herzog, J.~Hsu, \emph{et~al.}, ``Rt-1: Robotics transformer for real-world control at scale,'' \emph{arXiv preprint arXiv:2212.06817}, 2022.

\bibitem{janner2021offline}
M.~Janner, Q.~Li, and S.~Levine, ``Offline reinforcement learning as one big sequence modeling problem,'' \emph{Advances in neural information processing systems}, vol.~34, pp. 1273--1286, 2021.

\bibitem{zeng2021transporter}
A.~Zeng, P.~Florence, J.~Tompson, S.~Welker, J.~Chien, M.~Attarian, T.~Armstrong, I.~Krasin, D.~Duong, V.~Sindhwani, \emph{et~al.}, ``Transporter networks: Rearranging the visual world for robotic manipulation,'' in \emph{Conference on Robot Learning}.\hskip 1em plus 0.5em minus 0.4em\relax PMLR, 2021, pp. 726--747.

\end{thebibliography}
}

\end{document}